\pdfoutput=1

\documentclass{article}

\usepackage[bookmarks=false]{hyperref}

\usepackage[nonatbib, final]{nips_2016_modified}
\usepackage[numbers]{natbib}
\usepackage[utf8]{inputenc} 
\usepackage[T1]{fontenc}    
\usepackage{url}            
\usepackage{booktabs}       
\usepackage{amsfonts}       
\usepackage{nicefrac}       
\usepackage{microtype}      

\usepackage{color}
\usepackage{graphicx}
\usepackage{subfigure}
\usepackage{bm}

\newcommand{\method}[1]{\textsc{#1}} 

\newcommand{\vct}[1]{\bm{#1}} 
\newcommand{\mat}[1]{\bm{#1}} 

\newcommand{\field}[1]{\mathbb{#1}}
\newcommand{\R}{\field{R}} 

\title{
Exploiting Convolutional Neural Network for Risk Prediction with
Medical Feature Embedding}

\author{
Zhengping Che \\
University of Southern California \\
\texttt{zche@usc.edu} \\
\And
Yu Cheng \\
IBM T. J. Watson Research Center \\
\texttt{chengyu@us.ibm.com} \\
\And
Zhaonan Sun \\
IBM T. J. Watson Research Center \\
\texttt{zsun@us.ibm.com} \\
\And
Yan Liu \\
University of Southern California \\
\texttt{yanliu.cs@usc.edu} \\
}

\begin{document}

\maketitle

\vspace{-0.25in}
\section{Introduction}
The widespread availability of electronic health records (EHRs)
promises to usher in the era of personalized medicine. However, the
problem of extracting useful clinical representations from
longitudinal EHR data, sometimes called computational phenotyping,
remains challenging, owing to the heterogeneous, longitudinally
irregular, noisy and incomplete nature of such data.

In this paper, our focus is on the problems of high dimensionality
and temporality. We explore deep neural network models with learned
medical feature embedding to deal with these issues. Specifically,
we use a multi-layer convolutional neural network (CNN) to parameterize
the model and is thus able to capture complex non-linear
longitudinal evolution of EHRs. Different from recent proposed deep
learning approaches such as stacked auto-encoders~\cite{vincent2010stacked} and recurrent neural network~\cite{hochreiter1997long,cho2014properties}, our model
can effectively capture local/short temporal dependency in EHRs,
which is beneficial for risk prediction. To account for high
dimensionality, instead of using the raw EHR data as the input, we
use the embedding medical features in the CNN model. Based on the
medical context, each medical event is compressed into a given
length vector with medical feature embedding. Similar to the word
embedding \cite{mikolov2013distributed}, the event embedding
presented in our model holds its natural medical concept.
Our initial experiments produce promising results, and demonstrate
the effectiveness of both the medical feature embedding and the
proposed convolutional neural network in risk prediction on cohorts
of congestive heart failure and diabetes patients, compared
with several strong baselines.

\section{Methodology}

\subsection{Medical Feature Embedding Learning}
\label{sec:method-w2v} We extend word2vec
model~\cite{mikolov2013distributed,Choi2016} to
learn medical feature embeddings from EHR data. Word2vec takes a
corpus of text sequences and learns the vector representations,
called embeddings, for each word. The words with similar contexts
usually have close embeddings in the vector space. In healthcare domain, Word2vec has also been used for medical texts and claims~\cite{choi2016learning}. 
Since our EHRs
are also sequential data with thousands of different medical events
(e.g., diagnosis, medications, lab test, etc.), it is natural to
generalize the ideas in word2vec to learn low-dimension and
meaningful embeddings for medical events given a large corpus of EHR
data. We take all the records for one patient as a sequence of
medical events. For each event in the sequence, the word2vec model
takes other events within a local window of current event into
consideration. We train word2vec with continuous bag-of-words (CBOW)
method.

\subsection{CNN Risk Prediction Model}

\begin{figure}[htb]
\setlength\fboxsep{0pt}
\centering
\includegraphics[width=0.61\columnwidth]{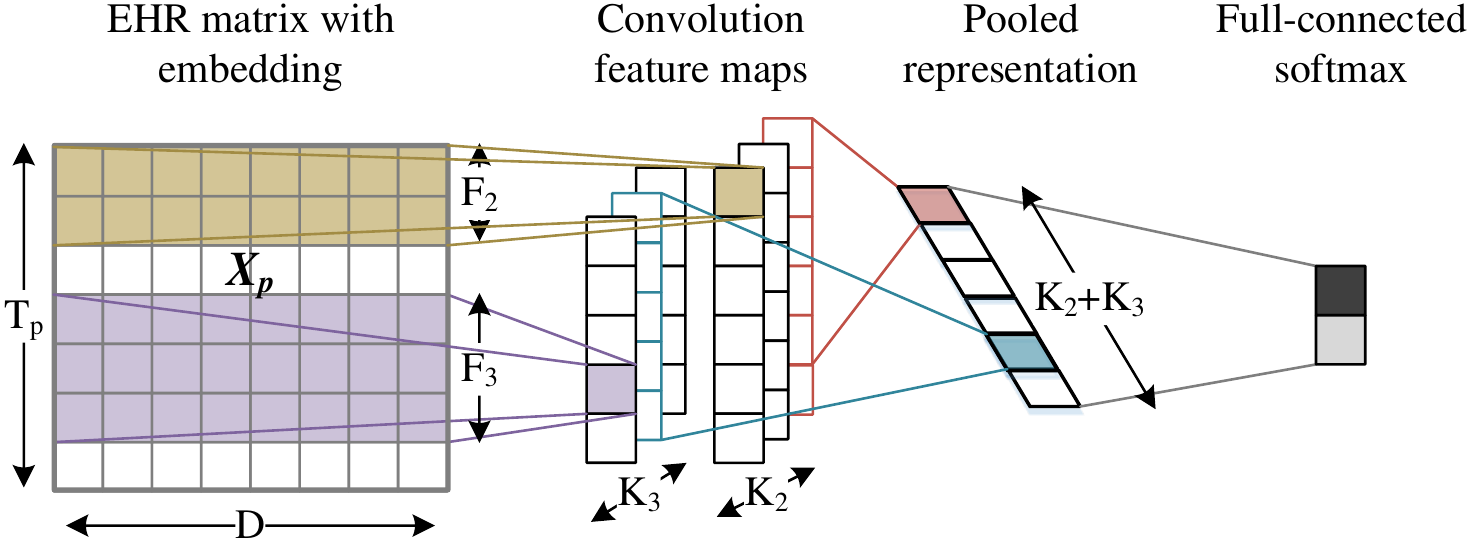}
\vspace{-0.05in}
\caption{Convolutional neural network prediction model (with filters of size 2 and 3).}
\vspace{-0.15in}
\label{fig:cnn-model}
\end{figure}

Convolutional neural network (CNN) is one type of neural network
models which is good at capturing local structure and has moderate
model size.
A variaity of CNN models are shown to be effective in
handling images~\cite{krizhevsky2012imagenet}, sequential
data~\cite{kalchbrenner2014convolutional}, and recently on medical lab tests~\cite{razavian2015temporal}.
In risk prediction tasks on EHR data of diagnoses and medications, the temporal structures and local dependencies across features also provide lots of useful information.
In order to effectively learn these useful characteristics from EHR data and make better predictions, we use a CNN with one-dimensional convolutional operation over the temporal dimension.
The input to our CNN model is the EHR records of patient $p$, which
is represented as a temporal embedding matrix $\mat{X}_p \in \R^{T_p
\times D}$, where $T_p$, which is usually different among the patients, is the number of medical events in patient
$p$'s record, and $D$
is the learned embedding dimension. Each row $\vct{x}_i \in \R^D,
i=1,\dots,T_p$ in $\mat{X}_p$ is the embedding vector of $i$th event
of that patient. Noticing that no spatial or temporal relationship
inherently exists along the embedding dimension in the EHR matrix,
we apply convolutional operation not over embedding dimension but
only over the temporal dimension. Assume that the CNN model has $K$
convolutional filters of size $F$, we will have $K$ different vector
outputs of length $T_p-F+1$.
A filter of size 2 captures
pairwise temporal dependency, while a filter of larger size might
capture longer temporal dependency.
That said, using a combination of filters with
different lengths benefits capturing dependencies in multiple levels
and thus improves the prediction, which is validated in our experiments.
After convolutional step, we apply
a max-pooling operation along the temporal dimensions to keep the
most important features across the time. This temporal pooling
produces output vector of length $K$ and moreover converts the
inputs with different temporal lengths into a fixed length vector
inherently. Finally a fully connected softmax layer is used to get
the prediction. An illustration of the proposed CNN prediction model
is shown in Figure~\ref{fig:cnn-model}. Our convolutional network
model is similar to the basic model shown in~\cite{chengrisk} but
with multiple scale filters. Furthermore, the input to our model is
temporal embedding matrix of events instead of raw event matrix. We
can also jointly fine-tune a task-specific embedding matrix together
with training the CNN model, or use both fixed and fine-tuned
embedding together as input matrix.

\section{Experiments}
\subsection{Experimental Setup}
\subsubsection{Medical Feature Embedding}
We applied medical feature embedding learning method described in Section~\ref{sec:method-w2v} from a private EHR dataset.
This dataset contains 218680 patients and 14969489 observations of 14690 unique medical events, which are ICD-9 diagnosis and medications. We only compute the embeddings for 8627 events that show at least 5 times in this dataset.
We set the local window size to be 20 for each event, which means at most 40 events nearby are considered to be the neighbors of that event.
We set the embedding dimension is 200 in all the following experiments, if not mentioned explicitly. Our medical feature embedding model is implemented based on python Gensim~\cite{rehurek_lrec} package.

\subsubsection{Risk Prediction}
We apply the convolutional neural network and the learned embeddings to predict two important diagnoses, diabetes and congestive heart failure, as two binary classification tasks separately.
We take the same private EHR dataset used in medical embedding learning, but only a small portion of patients in the dataset have the target diagnoses.
In order to handle this class imbalance issue and compare all methods robustly, we respectively extract subsets for these two diagnosis.
First, all patients with the target diagnosis are selected to build the case group.
We take both the ICD-9 diagnoses and medications as input features for heart failure prediction, but only diagnoses for diabetes since some medications may apparently indicate diabetes. If there are multiple events at the same time, we generate the input sequence by ordering them ascendingly by the event index.
As patients usually have medical records in different lengths, it might impact the prediction performance and prevent us from learning robust model and comparing models consistently.
We remove all patients with less than 50 medical event records, and keep the first up to 250 events before the time they get their first target diagnosis.
For each patient in the case group, two other patients without the target diagnosis but with similar demographic information and EHR record length are selected to form a control group. Similarly we limit the length of each record sequence between 50 and 250.
Finally we have 2248 and 4496 patients in case and control group for diabetes, and 3357 and 6714 patients in case and control group for heart failure.
We split all the data into training, validation, and test subsets by 7:1:2.
The CNN model we used in experiments has one convolutional layer with 100 filters of size 3, 4 and 5, followed by one pooling layer and one fully connected prediction layer. We take tanh function as activation in convolutional and fully connected layer since it performed better in our experiments.
We train the network using AdaDelta~\cite{zeiler2012adadelta} with default settings.

\subsection{Experimental Results}

In our experiments, we take classification accuracy (Accuracy), the area under receiver operating characteristic curve (AUROC), the area under precision-recall curve (AUPRC), and maximum F1 score (Max F1) to compare the performance of all methods in two risk prediction tasks.

\begin{table}[bt]
\scriptsize
\centering
\caption{Prediction performance comparison of convolutional neural network and baselines.}
\label{tab:main}
\begin{tabular}{llcccc}
\toprule
{\textbf{Method}} & {\textbf{Input}} & \multicolumn{2}{c}{\textbf{Heart Failure}} & \multicolumn{2}{c}{\textbf{Diabetes}} \\ \cmidrule{3-6}
 &  & \textbf{Accuracy} & \textbf{AUROC} & \textbf{Accuracy} & \textbf{AUROC} \\ \midrule \midrule
\method{CNN} & \method{W2v} & $0.8630$ & $\mathbf{0.9329}$ & $\mathbf{0.9844}$ & $\mathbf{0.9989}$ \\ \midrule
\method{CNN} & \method{Rand} & $0.8337$ & $0.8999$ & $0.9815$ & $0.9959$ \\ \midrule
\method{CNN} & \method{Raw} & $0.8511$ & $0.9203$ & $0.9748$ & $0.9953$ \\ \toprule
\method{LR} & \method{BofW} & $0.8094$ & $0.8652$ & $0.9066$ & $0.9681$ \\ \midrule
\method{SVM} & \method{BofW} & $0.7906$ & $0.8397$ & $0.9044$ & $0.9687$ \\ \midrule
\method{RF} & \method{BofW} & $0.8620$ & $0.9311$ & $0.9029$ & $0.9683$ \\ \midrule
\method{LR} & \method{W2v} & $0.8625$ & $0.9289$ & $0.9266$ & $0.9802$ \\ \midrule
\method{SVM} & \method{W2v} & $0.8476$ & $0.9140$ & $0.9148$ & $0.9753$ \\ \midrule
\method{RF} & \method{W2v} & $\mathbf{0.8650}$ & $0.9269$ & $0.8955$ & $0.9654$ \\ \bottomrule
\end{tabular}
\vspace{-0.1in}
\end{table}

\subsubsection{Risk Prediction Comparison}
\label{sec:exp-pred}
First we want to show the performance of our proposed predict model.
We compare logistic regression (\method{LR}), linear support vector machine (\method{SVM}) and random forest (\method{RF}) as baselines.
We apply L2 regularizers in \method{LR} and \method{SVM}.
We use early stopping for \method{RF} with at most 50 trees.
We evaluate baselines with either raw inputs or learned embeddings.
For raw input (\method{BofW}), we take bag of words to keep only the frequencies of all events, thus the input vector is of length 8627.
For word2vec embedding input (\method{W2v}), we take the embeddings of all events in that record.
In order to make the patient records in different temporal lengths to be the same length. we take the columnwise aggregation values along the temporal dimension, including summation, maximum, and minimum of all embeddings in the record, which is equivalent to \method{W2v-All} shown in Section~\ref{sec:exp-w2v}. In this way, the input length is related to only the embedding dimension and much smaller than raw input length.
For CNN model, besides the learnt medical feature embeddings (\method{W2v}), we also evaluate the performance with raw feature index as input(\method{Raw}), and use random vector to initialize embedding and jointly trained embeddings together with CNN model (\method{Rand}).
Table~\ref{tab:main} shows the classification accuracy and AUROC on the two prediction tasks.
The proposed \method{CNN} methods with learned embeddings is among the best methods in heart failure prediction task and significantly outperforms baselines in diabetes prediction task.
The performance improvement mainly comes from the learned embeddings for heart failure task, but from CNN model structures for diabetes task.
The learned embedding helps \method{LR} and \method{SVM} a lot, but \method{RF} barely benefits from the it, which is probably because \method{RF} selects discrete input feature at each tree node.

\begin{table}[tb]
\scriptsize
\centering
\caption{Learned embeddings v.s. raw inputs on logistic regression prediction model.}
\label{tab:w2v-results}
\begin{tabular}{lcccc|cccc}
\toprule
 & \multicolumn{4}{c}{\textbf{Heart Failure}} & \multicolumn{4}{c}{\textbf{Diabetes}} \\ \cmidrule{2-9}
 & \textbf{Accuracy} & \textbf{AUROC} & \textbf{AUPRC} & \textbf{Max F1} & \textbf{Accuracy} & \textbf{AUROC} & \textbf{AUPRC} & \textbf{Max F1} \\ \midrule \midrule
\method{BofW} & $0.8094$ & $0.8652$ & $0.7590$ & $0.7302$ & $0.9066$ & $0.9681$ & $0.9386$ & $0.8654$ \\
 \midrule \method{Rand-Sum} & $0.8005$ & $0.8784$ & $0.7818$ & $0.7382$ & $0.8488$ & $0.9317$ & $0.8869$ & $0.8030$ \\
\toprule
\method{W2v-Ave} & $0.8471$ & $0.9119$ & $0.8133$ & $0.7792$ & $0.8977$ & $0.9583$ & $0.8848$ & $0.8639$ \\ \midrule
\method{W2v-Sum} & $0.8407$ & $0.9142$ & $0.8389$ & $0.7797$ & $\mathbf{0.9266}$ & $ \mathbf{0.9802}$ & $\mathbf{0.9638}$ & $\mathbf{0.8950}$ \\ \midrule
\method{W2v-Max} & $0.8422$ & $0.9157$ & $0.8267$ & $0.7878$ & $0.8695$ & $ 0.9431$ & $0.8654$ & $0.8168$ \\ \midrule
\method{W2v-All} & $\mathbf{0.8625}$ & $\mathbf{0.9289}$ & $\mathbf{0.8668}$ & $\mathbf{0.8056}$ & $0.9162$ & $ 0.9740$ & $0.9530$ & $0.8763$ \\ \bottomrule
\end{tabular}
\end{table}

\subsubsection{Feature Embedding Evaluation}
\label{sec:exp-w2v}
Next, we compare the learned medical feature embeddings and raw representations by logistic regression prediction models.
Comparing with raw feature index \method{BofW}, we take several representations based on learned embeddings:
\method{W2v-Ave}, \method{W2v-Sum},\method{W2v-Max} takes the average, summation, and maximum values of the columnwise embedding inputs, respectively. The input size is 200 for these three methods. We also take a combination of summation, minimum and maximum values, which is used in Section~\ref{sec:exp-pred}, as \method{W2v-All}. Additionally, we take the summation of random embeddings (\method{Rand-Sum}) as comparison.
As shown in Table~\ref{tab:w2v-results}, our learned word2vec embeddings outperform the raw input and random embedding, especially for heart failure prediction.

\subsubsection{Early Prediction}
Making better prediction in advance can help doctor make timely decisions and diagnose diseases in their early stage.
We also test our model in a simulated early stopping setting. For each patient in the case group, we only take the observations before a period of time from the time that the first target diagnosis is confirmed to the patient.
We set the hold-off period to be 90/180 days in our experiments, to predict whether the patient will have the target diagnosis in 90/180 days later.
Some of the patients do not have enough records before the first target diagnosis and thus are removed from the dataset. Thus the longer hold-off period we select, the fewer patients will be taken in the case group which makes the problem more difficult. We also remove some patients in control group to make sure the size of control group is always double of that of case group.
The number of patients and early prediction performance is shown Table~\ref{tab:early-pred}. As expected, the performance decreases when the duration goes up, but it's still comparable with baselines shown in Table~\ref{tab:main}. For instance, our 180-day early prediction on diabetes achieves Accuracy of 0.9277, which is even higher than the that of best baseline with full observation, which accuracy of 0.9266 from \method{LR+W2v-All}.

\begin{table}[b!]
\scriptsize
\centering
\caption{Early prediction results for CNN prediction model.}
\label{tab:early-pred}
\begin{tabular}{lccc|ccc}
\toprule
 & \multicolumn{3}{c}{\textbf{Heart Failure}} & \multicolumn{3}{c}{\textbf{Diabetes}} \\ \cmidrule{2-7}
 & \# of Case & Accuracy & AUROC & \# of Case & Accuracy & AUROC \\ \midrule
0 days & $3357$ & $0.8630$ & $0.9329$ & $2248$ & $0.9844$ & $0.9989$ \\ \midrule
90 days & $2573$ & $0.8329$ & $0.8889$ & $1616$ & $0.9835$ & $0.9988$ \\ \midrule
180 days & $2105$ & $0.8274$ & $0.8995$ & $1221$ & $0.9277$ & $0.9716$ \\ \bottomrule
\end{tabular}
\vspace{-0.1in}
\end{table}

\section{Discussion}
In this work, we proposed a general framework for risk prediction tasks by convolutional neural network and learned medical feature embedding on EHR data.
We found that the learned embeddings of are better than raw representations in both baselines and our proposed CNN models.
We also evaluated our framework in details and demonstrated its capability to capture complex temporal structures and make reasonable early predictions. In the future we plan to consider temporal gaps of events in embedding and prediction model, and find clinical interpretations of the learned embeddings and filters in the network.

\small{
    \bibliographystyle{abbrv}
    \bibliography{references}
}
\end{document}